 \documentclass[final,5p,times,twocolumn,authoryear]{elsarticle}

\usepackage{lipsum}
\usepackage{amssymb}
\usepackage{amsmath}
\usepackage{tabularx}
\usepackage{booktabs}
\usepackage{color, colortbl}
\usepackage{graphicx}
\usepackage{caption}
\usepackage{multirow}
\usepackage{subcaption}

\journal{}

\begin{document}

\begin{frontmatter}



\title{Improving Facade Parsing with Vision Transformers and Line Integration}


\author[first]{Bowen Wang}
\ead{wang@ids.osaka-u.ac.jp}
\author[label2]{Jiaxin Zhang}
\ead{Corresponding: jiaxin.arch@ncu.edu.cn}
\author[label3]{Ran Zhang}
\ead{reeseran1217@gmail.com}
\author[label2]{Yunqin Li}
\ead{liyunqin@ncu.edu.cn}
\author[label1]{Liangzhi Li}
\ead{li@ids.osaka-u.ac.jp}
\author[label1]{Yuta Nakashima}
\ead{n-yuta@ids.osaka-u.ac.jp}

\affiliation[first]{organization={Science and Technology, Graduate School of Information, Osaka University},
            addressline={2-1,Yamadaoka}, 
            city={Osaka},
            postcode={5650871}, 
            state={Suita},
            country={Japan}}

\affiliation[label2]{organization={Architecture and design college, Nanchang University},
            addressline={No. 999, Xuefu Avenue, Honggutan New District}, 
            city={Nanchang},
            postcode={330031}, 
            country={China},
            }

\affiliation[label3]{organization={Faculty of Foreign Studies,Beijing Language and Culture University},
            addressline={15 Xueyuan Road, Haidian District}, 
            city={Beijing},
            postcode={100083}, 
            country={China},
            }

\begin{abstract}
Facade parsing stands as a pivotal computer vision task with far-reaching applications in areas like architecture, urban planning, and energy efficiency. Despite the recent success of deep learning-based methods in yielding impressive results on certain open-source datasets, their viability for real-world applications remains uncertain. Real-world scenarios are considerably more intricate, demanding greater computational efficiency. Existing datasets often fall short in representing these settings, and previous methods frequently rely on extra detection models to enhance accuracy, which requires much computation cost. In this paper, we first introduce Comprehensive Facade Parsing (CFP), a dataset meticulously designed to encompass the intricacies of real-world facade parsing tasks. Comprising a total of 602 high-resolution street-view images, this dataset captures a diverse array of challenging scenarios, including sloping angles and densely clustered buildings, with painstakingly curated annotations for each image. We then propose a new pipeline known as Revision-based Transformer Facade Parsing (RTFP). This marks the pioneering utilization of Vision Transformers (ViT) in facade parsing, and our experimental results definitively substantiate its merit. We also design Line Acquisition, Filtering, and Revision (LAFR), an efficient yet accurate revision algorithm that can improve the segment result solely from simple line detection using prior knowledge of the facade. In ECP 2011, RueMonge 2014, and our CFP, we evaluate the superiority of our method. 
\end{abstract}




\end{frontmatter}



\section{Introduction}
With the burgeoning demand for 3D architectural models in digital twin cities, autonomous driving, and urban simulations \cite{bagloee2016autonomous,huang2023bim}, facade parsing—particularly detailed parsing of windows and doors in CityGML Level of Detail 3 (LoD3) architectural models \cite{donkers2016automatic}—has become paramount in urban 3D reconstruction. Prevailing facade parsing approaches predominantly lean on syntactical rules \cite{eilouti2019shape} or rudimentary computer vision techniques \cite{liu2020deepfacade,zhang2021automatic}. Yet, these methods grapple with challenges. Syntactical rules, typically mined from architectural design tenets, struggle to encapsulate the broad heterogeneity of architectural styles, leading to potential parsing incompleteness. Additionally, fundamental computer vision techniques like region growing and edge detection, contingent upon local gradients or localized intensity variances, exhibit noise susceptibility, thereby undermining image analyses' stability and accuracy.

\begin{figure}[t]
    \centering
    \begin{subfigure}[b]{0.36\columnwidth}
        \includegraphics[width=\columnwidth]{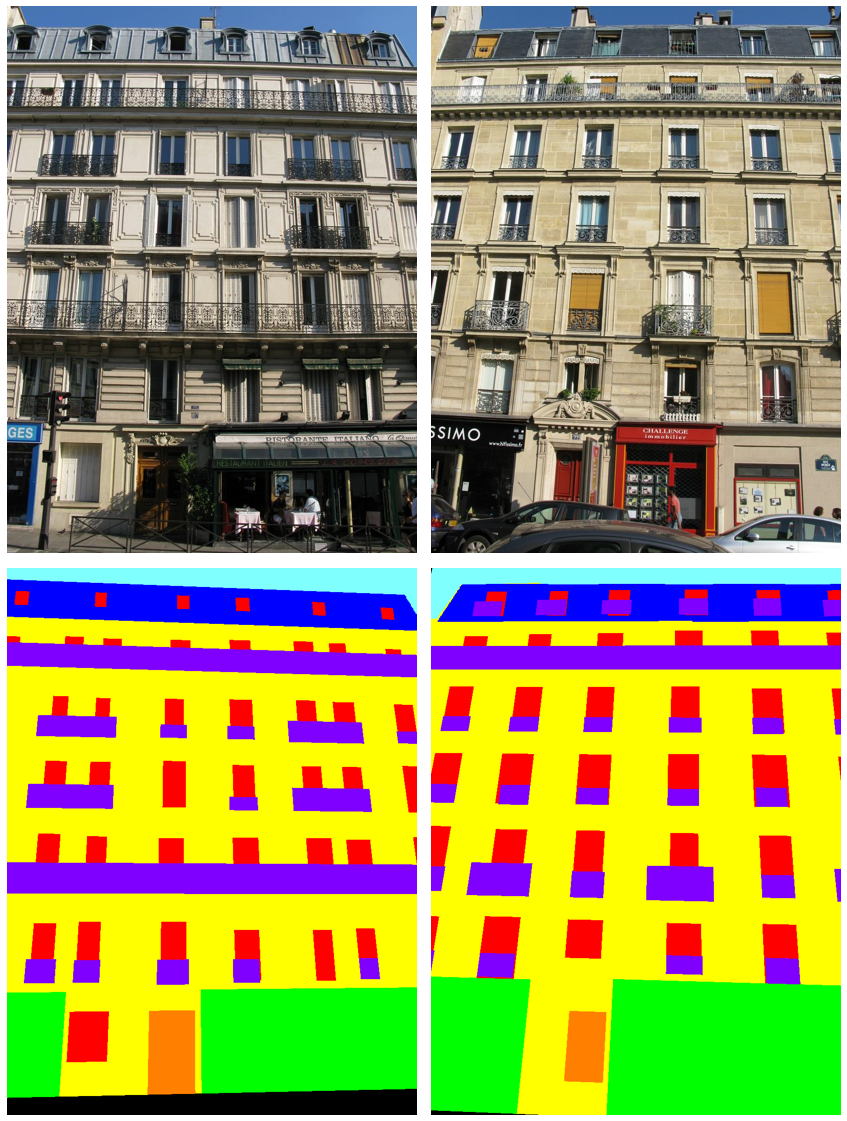}
        \caption{RueMonge 2014.}
        \label{fig:ruemonge}
    \end{subfigure}
    \begin{subfigure}[b]{0.62\columnwidth}
        \includegraphics[width=\columnwidth]{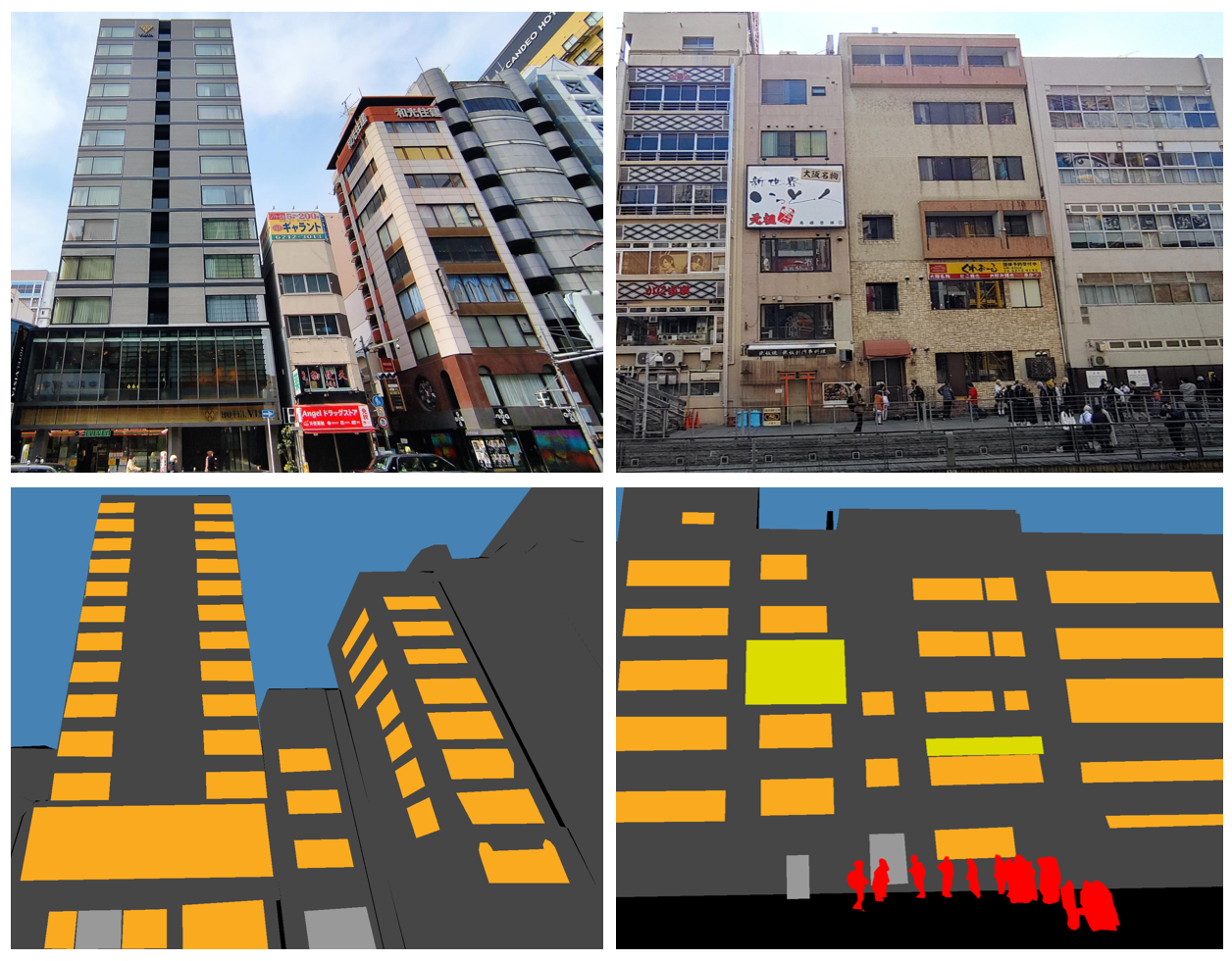}
        \caption{Our CFP Dataset.}
        \label{fig:ours}
    \end{subfigure}
    \begin{subfigure}[b]{0.93\columnwidth}
        \includegraphics[width=\columnwidth]{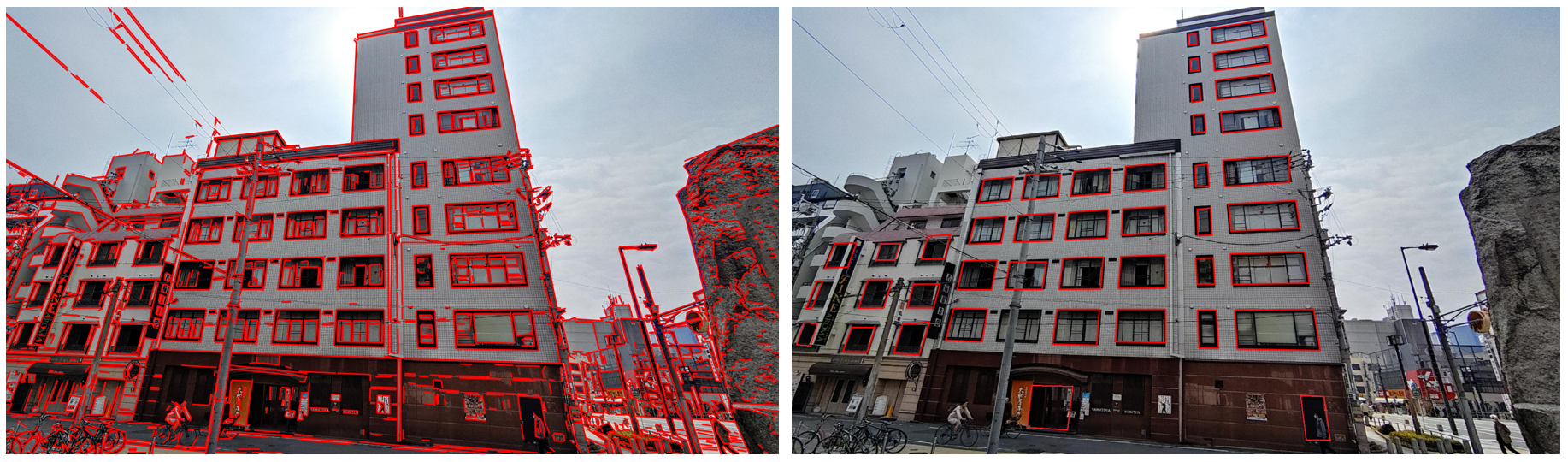}
        \caption{Inference sample by LAFR.}
        \label{fig:LARF}
    \end{subfigure}
    \caption{Image samples from (a) RueMonge 2014 and (b) our CFP dataset. In Figure (c), we show an inference sample by our LARF (left is the line detection result from LSD and right is the integrated lines after LARF). It has the ability to outline facade elements, e.g., windows and doors, through simple line detection.
    }
    \label{fig:kinetics-visualizations}
    \vspace{-0.1in}
\end{figure}

Recent advancements in deep learning proffer enhanced insights into image comprehension \cite{lecun2015deep}. Convolutional Neural Networks (CNNs) excel in discerning intricate hierarchical features within images and consistently attain state-of-the-art (SOTA) results across diverse domains \cite{li2021survey,gu2018recent}. Antecedent studies \cite{femiani2018facade,ma2020pyramid} employing CNNs for semantic facade parsing have outperformed traditional methods on open-source datasets like eTRIMs \cite{korc2009etrims}, ECP2011 \cite{teboul2010ecole}, Graz2012 \cite{riemenschneider2012irregular}, and CMP2013 \cite{tylevcek2013spatial}. Certain investigations \cite{liu2017deepfacade,DAI2021107921} harness facade priors to augment segmentation quality, postulating rectangularity of facade elements (e.g., windows) and achieving this end through object detection \cite{girshick2015fast}. Although such revisions can benefit to segment results, one defect is that they require high extra computation costs. 

Besides, existing methodologies are tailored for extant datasets characterized by front-facing architectural views, controlled illumination, and minimal occlusions \cite{rohlig2017visibility}. Nevertheless, these datasets' volume and architectural style diversity fall short of meeting the intricate demands of contemporary deep learning architectures. The predominance of rectified images in most datasets suggests potential performance bottlenecks in real-world applications. The constraints in image resolutions of datasets also hint at suboptimal generalization capabilities in realistic scenarios.

Furthermore, previous methodologies for facade parsing utilizing CNN-based approaches were limited, primarily due to the inherent inability of rudimentary neural networks to model long-range pixel interactions, thereby compromising optimal feature representation through contextual information. Recently, the Vision Transformer (ViT) \cite{dosovitskiy2020image}, a revolutionary deep learning architecture, has heralded notable advancements within the realm of computer vision \cite{strudel2021segmenter,khan2022transformers}. Intrinsically designed to discern contextual relationships and adeptly handle high-resolution imagery through segmented patches, it emerges as an exemplar for semantic segmentation tasks \cite{zhao2017pyramid,minaee2021image}. However, in spite of its evident potential, the exploration of ViT in facade segmentation remains embryonic, attributed largely to its voracious data appetites \cite{steiner2021train} and the prevailing paucity of comprehensive facade datasets.


In summary, the primary challenges previously encountered in facade parsing can be distilled into three focal areas: (1) Bridging the discrepancy between existing facade segmentation datasets and real-world scenarios to enhance the robustness of facade segmentation techniques; (2) Harnessing the rich contextual dependencies inherent in local features to achieve superior feature representation; and (3) Constructing a revision algorithm grounded in prior knowledge, synergizing the intricate hierarchical feature extraction capabilities of deep learning, to bolster the precision and efficiency of facade image parsing in authentic scenarios.

\begin{table*}[ht]
\caption{Introduction of previous facade datasets and our CFP Datasets}
\centering
\resizebox{0.95\linewidth}{!}{
\begin{tabular}{p{4cm} p{7.5cm} p{3.2cm} p{2cm} p{3.5cm}}
\toprule
\textbf{Dataset} & \textbf{Description} & \textbf{Number of Images} & \textbf{Resolution} & \textbf{Other Information}\\
\midrule
ECP 2011 \cite{teboul2010ecole} & Labeled pictures in Paris - 7 categories: walls, windows, doors, balconies, roofs, stores, sky & 104 & 1024x768 & -\\
\\
Graz 2012 \cite{riemenschneider2012irregular} & Images from Germany and Austria - 4 categories: door, window, wall, sky & 50 & 640x480 & -\\
\\
CMP 2013 \cite{tylevcek2013spatial} & Facade images from around the world - 12 categories: facade, molding, cornice, pillar, window, door, sill, blind, balcony, shop, deco, background & 378 basic + 228 extended & 512x512 & -\\
\\
ENPC 2014 \cite{lotte20183d} & Images located in Paris - Same categories as ECP2011 & 79 & 1280x960 & -\\
\\
RueMonge 2014 \cite{riemenschneider2014learning} & 428 multi-view images of a street in Paris. It provides annotations with seven semantic categories for 2D images, including door, shop, balcony, window, wall, roof, and sky. & 428 images (60 buildings) & 800x1067 & Dataset for 3D reconstruction and semantic mesh annotation.\\
\\
eTRIMs \cite{korc2009etrims} & Multi-view images based on multiple European cities - 8 categories: building, car, door, pavement, road, sky, vegetation, window & 60 & 960x720 & Includes three sets of annotations for object segmentation, class segmentation, and object boundaries.\\
\\
LabelMeFacade \cite{Froehlich-Rodner-Denzler-ICPR2010} & Based on the eTIRMs extended LabelMe database - 8 categories: building, car, door, pavement, road, sky, vegetation, window & 945 & 256x256 & Only pixel-level masks are provided.\\
\midrule
CFP (\textit{Ours}) & Images are from six cities and captured from different angles. There are 9 classes: building, window, door, roof, tree, sky, people, car, and sign.  & 602 & 2560×1440 & Provide instance-level mask annotation for all objects.\\
\bottomrule
\end{tabular}
}
\label{dataset_}
\end{table*}

\begin{figure*}[t]
    \centering
    \begin{subfigure}[b]{1\columnwidth}
        \includegraphics[width=\columnwidth]{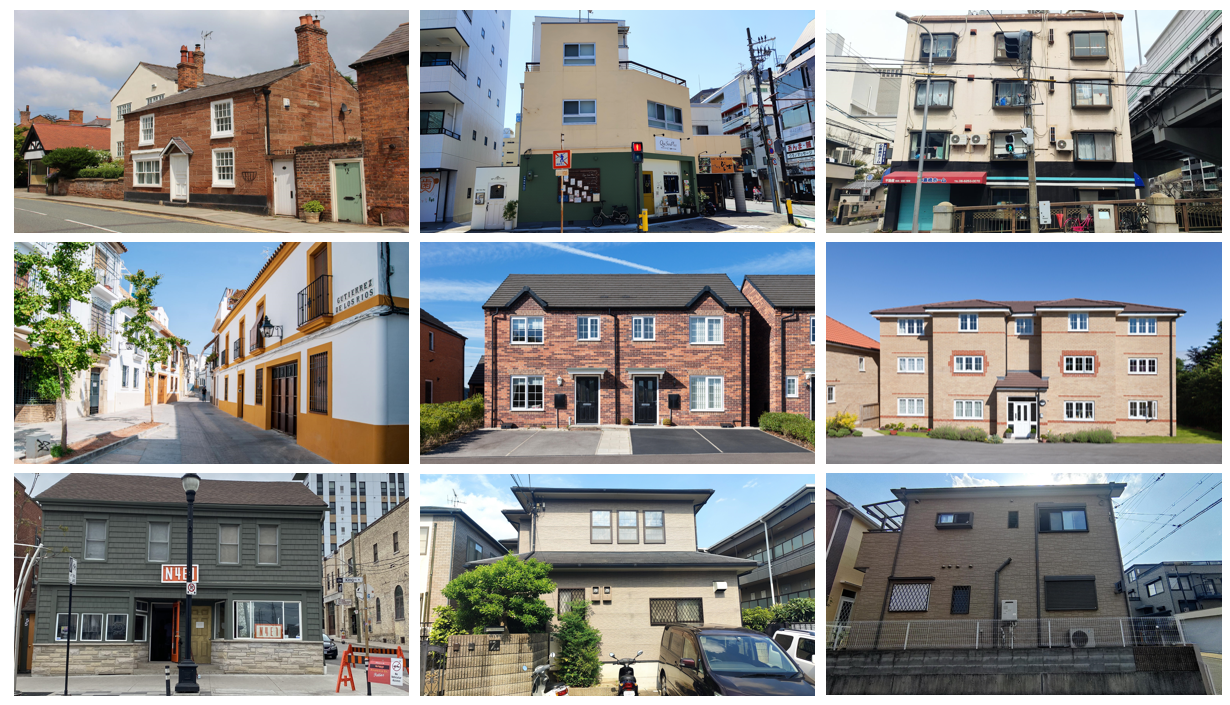}
        \caption{Samples from residential area.}
        \label{fig:sample1}
    \end{subfigure}
    \begin{subfigure}[b]{1\columnwidth}
        \includegraphics[width=\columnwidth]{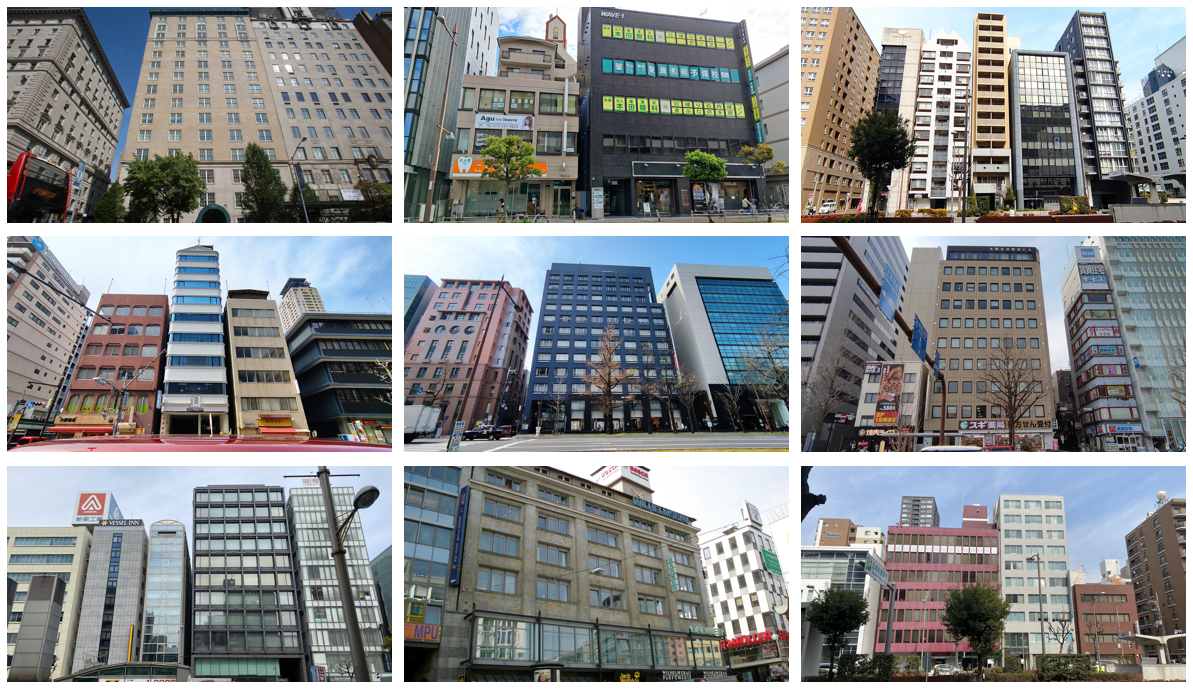}
        \caption{Samples from city center.}
        \label{fig:sample2}
    \end{subfigure}
    \caption{Some image samples from our CFP dataset.}
    \label{fig:more_samples}
\end{figure*}

In this paper, we first introduce a novel streetscape image dataset, designated Comprehensive Facade Parsing (CFP), to address these disparities, tailored for facade parsing. Diverging from extant datasets, ours encompasses annotated architectural imagery from six cities—Osaka, Tokyo, Toronto, Shanghai, Nanjing, and Nanchang—offering diverse perspectives, intricate lighting conditions, foreground occlusions, and a tapestry of architectural styles. Both semantic segmentation and object detection annotations are rendered for facade components. Samples comparison between RueMonge and CFP are shown in Figure \ref{fig:ruemonge} and Figure \ref{fig:ours}. Images in our dataset are more challenging. Furthermore, we propose a pioneering methodology, Revision-based Transformer Facade Parsing (RTFP), optimized for streetscape imagery. This paradigm pivots on a semantic segmentation model grounded in the Vision Transformer (ViT) \cite{dosovitskiy2020image}, geared towards preliminary segmentation. The ViT demonstrates superior proficiency in discerning global layouts, which is pivotal for precise facade parsing, compared to preceding CNN-centric models. Additionally, we incorporate Masked Autoencoders (MAE) \cite{he2022masked}, a self-supervised pre-training algorithm, to foster enhanced fine-tuning of the ViT \cite{dosovitskiy2020image} on facade-centric data. Conclusively, we unveil a Line Acquisition, Filtering, and Revision (LAFR) algorithm dedicated to refining rules for facade elements like windows (shown in Figure \ref{fig:LARF}). The LAFR emerges as an efficient and precise facade element prediction refinement instrument. Our LAFR relies on rudimentary line detection and artificially set refinement stipulations, eschewing heavyweight detection models \cite{girshick2015fast,he2017mask}.

\textbf{Contribution}. In this study, we introduced a novel facade parsing dataset CFP, which serves as a comprehensive testbed for evaluating the performance of contemporary facade segmentation techniques in real-world scenarios. Encompassing a wide array of street-view facade scenarios, CFP presents a challenging benchmark that pushes the boundaries of research in this domain. To facilitate precise and efficient facade parsing, we developed the ViT-based RTFP framework, which represents a significant contribution to the field. Notably, we introduced MAE pre-training specifically tailored for facade segmentation, marking the first application of this approach in the context of facade-related tasks. Furthermore, we introduced a straightforward yet powerful revision method called LAFR, designed to enhance facade predictions. Our experimental results unequivocally establish RTFP's superiority over previous methods across multiple datasets, underscoring its potential as a valuable tool in the arsenal of facade segmentation techniques.

\section{Related Works}\label{sec:relate}

\subsection{Datasets for Facade Parsing}
Facade parsing has emerged as a pivotal domain in architectural analysis and urban design, gaining significant attention in the recent computational vision literature. Table 1 shows the previous and our proposed facade parsing datasets. However, a comprehensive analysis of existing datasets reveals a range of challenges that remain unaddressed.

Historically, initial datasets often suffered from limited volume, containing mere hundreds to thousands of images, which impedes the training of sophisticated deep learning models \cite{mahajan2018exploring}. The diversity of these datasets has been another recurrent issue. Predominantly centered on specific architectural styles, regions, or types, many datasets exhibit a narrow scope, potentially compromising the model's adaptability to diverse facade designs \cite{jeong2019complex}. The intricacies of manual annotating have led to inaccuracies, especially in datasets dealing with multifaceted facade structures \cite{huang2020caption}. Moreover, a holistic representation is often neglected with primary emphasis on larger structures, overlooking finer elements like windows, doors, or balconies \cite{rohrbach2016grounding}. Real-time contextual information, crucial for practical applications, is often absent. A significant portion of these datasets predominantly captures imagery from fixed angles and specific conditions, neglecting the variability encountered in real-world scenarios \cite{neuhold2017mapillary}.

In the pursuit of curating an optimal facade parsing dataset, a few guidelines emerge from the pitfalls of predecessors. Ensuring diversity by capturing images from varied styles, regions, and lighting conditions can significantly bolster model generalizability \cite{hudson2019gqa}. Given the demands of deep learning, collecting extensive data, potentially in the magnitude of hundreds of thousands of images, becomes indispensable \cite{gupta2019deep}. A layered, meticulous annotation, capturing both macro and microelements, promises a richer dataset \cite{zhang2022automatic}. Embracing semi-automatic annotation techniques can expedite the process while retaining accuracy \cite{anisetti2017semi}. Capturing images across scales and angles, and harnessing high-resolution equipment can further enrich the data \cite{kattenborn2020convolutional}.

To address the disparity between existing facade segmentation datasets and real-world scenarios, as previously discussed, we introduce the Comprehensive Facade Parsing (CFP) dataset as a pioneering benchmark. As illustrated in Figure \ref{fig:more_samples}, the CFP dataset encompasses diverse street views, spanning residential areas to the city center, thereby providing a holistic representation of real-world scenarios.

\subsection{Deep Learning Methods for Facade Parsing}
Over the years, the landscape of facade segmentation has been profoundly transformed by the remarkable strides made in deep learning techniques, culminating in state-of-the-art (SOTA) achievements and substantial progress in the field. In this section, we delve into critical studies that have harnessed the power of deep learning methods to advance facade segmentation.

Some pioneering endeavors, such as those by Cohen et al. \cite{cohen2014efficient}, Mathias et al. \cite{mathias2016atlas}, and Kelly et al. \cite{kelly2017bigsur}, embraced the early stages of deep learning methods \cite{long2015fully,ronneberger2015u} for facade segmentation. These studies relied on relatively straightforward network architectures and robust yet heavy backbones. While they successfully achieved facade parsing, their performance left room for improvement.

Recent research has witnessed a shift towards designing specialized structures within deep learning models to enhance facade segmentation. For instance, Femiani et al. \cite{femiani2018facade} introduced the compatibility network, which concurrently addresses segmentation across various facade element types. In a similar vein, ALKNet \cite{ma2020pyramid} devised a pyramid structure equipped with ALK modules to obtain dependencies with a long-range among building elements across multiscale feature maps. Notably, these approaches outperformed the earlier pioneer works, showcasing significant advancements in facade segmentation. Nevertheless, it's essential to highlight that they did not incorporate prior knowledge about the construction of building facades into their methodologies.

Another aspect of facade parsing involves leveraging prior knowledge to enhance segmentation results. DeepFacade \cite{liu2017deepfacade} introduced object detection for window revision. They implemented Faster-RCNN \cite{ren2015faster} to automatically identify window objects and utilized their bounding boxes to refine segmentation. This innovative approach has inspired the development of several subsequent methods. FacMagNet \cite{DAI2021107921}, for instance, employs Mask-RCNN \cite{he2017mask} to segment individual facade elements, enhancing segmentation accuracy. Additionally, a side branch is incorporated to further improve the results, particularly for smaller objects like windows. Kong et al. \cite{kong2020enhanced} introduced the use of YOLO \cite{redmon2016you} for facade object detection, prioritizing efficiency. Zhang et al. \cite{zhang2022pr} adopted the DETR \cite{carion2020end} model to achieve superior detection and semantic segmentation performance. However, it's crucial to note that a common challenge with these methods lies in their reliance on large object detection models, which may present practical limitations in real-world applications due to computational demands and resource constraints. There is also no architecture based on ViT, thus it is of great interest to explore the usage of ViT for facade parsing.

\begin{figure*}[!t]
	\centering
	\includegraphics[width=0.90\textwidth]{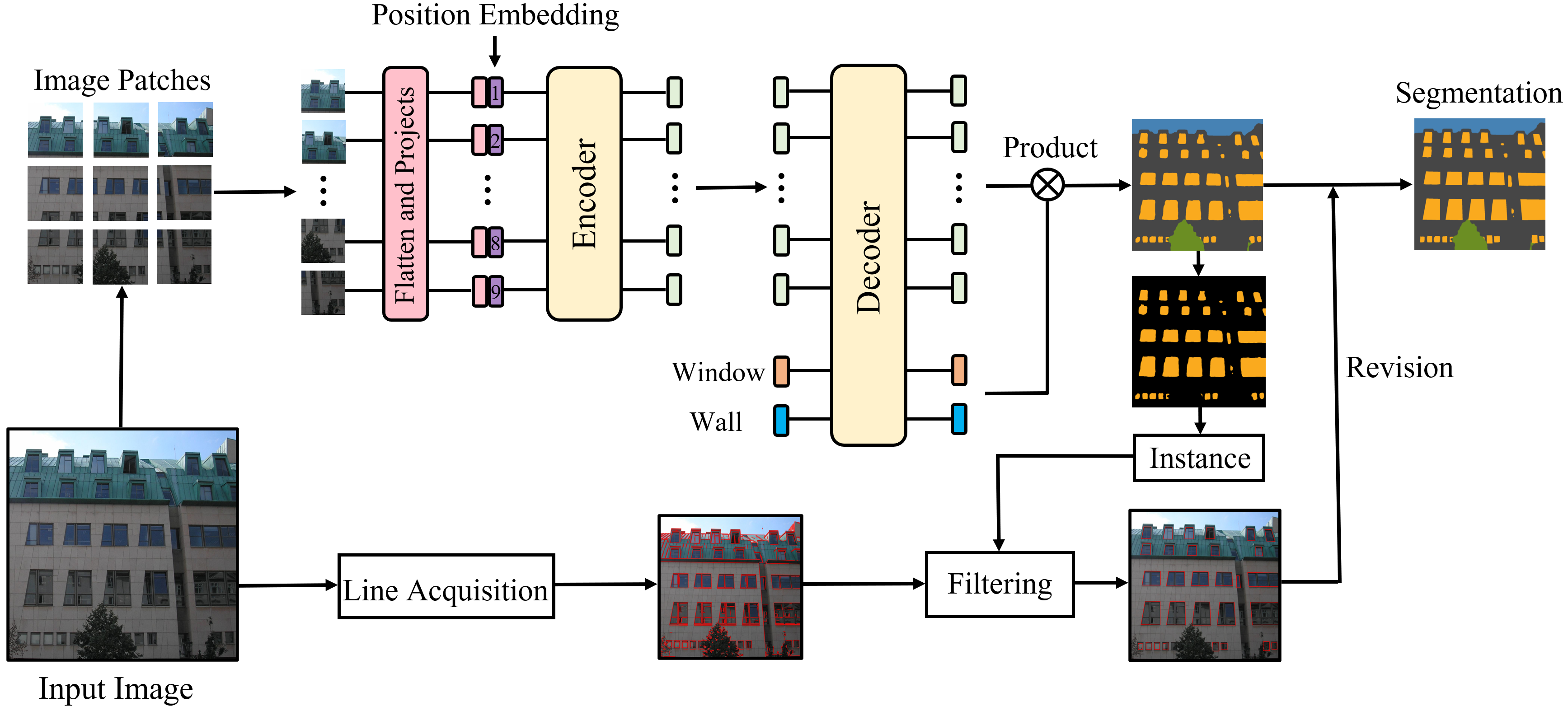}
	\caption{The pipeline of our Rvision-based Transformer Facade Parsing (RTFP). It is composed of two branches: (a) On the upper part, a ViT-based semantic segmentation model. (b) Our proposed method for line acquisition, filtering, and prediction revision is on the lower part.}
	\label{overview}
	\vspace{+0.1in}
\end{figure*}

\section{Method}\label{sec:method}
\subsection{Overview}
As shown in Figure \ref{overview}, our facade segmentation pipeline consists of two branches. In the upper branch, the input image is segmented into image patches and we feed them to our ViT based semantic segmentation model, which produces a preliminary prediction of the facade semantics. In the lower branch, we use traditional line detection methods to find lines that describe the outline of the facade elements. We then use each element instance from the preliminary prediction to filter the lines, keeping only those that perfectly frame out each element and integrating a corresponding outline. Finally, we revise the preliminary prediction using integrated outlines and produce the final segmentation map.

\subsection{ViT-based Semantic Segmentation}
Our ViT-based semantic segmentation is based on the architecture of Segmenter \cite{strudel2021segmenter}, which features a fully transformer-based encoder-decoder structure. This architecture maps the embeddings of image patches to pixel-level class predictions and comprises two main components: the encoder, which extracts information from raw image patches, and the decoder, which produces the final output. In the following sections, we will introduce them in detail. 

\subsubsection{Encoder}
We adopted an encoder to extract features from input image $x \in \mathbb{R}^{H \times W \times C}$. $x$ will be first divide into image patches $x = [x_1, ..., x_N] \in \mathbb{R}^{N \times P^2 \times C}$. $P$ is the patch size and $N$ is the number of patches calculated by $N=WH/P^2$. As usual ViT process, to create a sequence of patch embedding $x_E = [E_{x_1}, ..., E_{x_N}] \in \mathbb{R}^{N \times D}$, $x$ is scanned by a 2D convolution with both kernel size and stride as $P$. This operation will project each patch into an embedding vector $E \in \mathbb{R}^{D}$. In order to encode positional information, the architecture uses learnable position embeddings $pos = [pos_1, ..., pos_N ] \in \mathbb{R}^{N \times D}$. We add these embeddings to the sequence of patches to generate the tokens for the input sequence, which is represented by the equation $s_0 = x_E + pos$.

The Encoder is a transformer-based \cite{vaswani2017attention} architecture that consists of $L$ layers and is used to produce contextualized encoding $s_L \in \mathbb{R}^{N \times D}$. This powerful structure has revolutionized natural language processing by enabling models to capture complex relationships between words and has been extended to the computer vision area \cite{dosovitskiy2020image}. Each layer of the Encoder applies a multi-headed self-attention (MSA) block and then computed by a point-wise MLP block to refine the representation of the input as:
\begin{align}
    a_{i-1} = MSA(LN(s_{i-1})) + s_{i-1}, \\
    s_i = MLP(LN(a_{i-1})) + a_{i-1},
\end{align}
where $i$ is the layer index of $L$ and $LN$ is the layer normalization. The self-attention is composed of three point-wise linear layers that map patches $s_i$ to intermediate representations, including queries $Q \in \mathbb{R}^{N \times D}$, keys $K \in \mathbb{R}^{N \times D}$, and values $V \in \mathbb{R}^{N \times D}$. The queries are used to score the relevance of the keys for each patch, and the values are weighted by these scores and integrated to generate the final contextualized representation. This process allows the model to capture complex relationships among patches, resulting in highly informative and expressive encoding. It is formulated as follows:
\begin{align}
    MSA(Q, K, V) = softmax(\frac{QK^T}{\sqrt{d}})V.
\end{align}

Input patch sequence $s_0$ is mapped into $s_L$ by the encoder, containing rich image semantic information. $s_L$ will be adopted in a decoder (introduced in the continuing section) for preliminary semantic prediction.

\subsubsection{Decoder}
Decoder is also a full transformer-based structure with $M$ layers. Following previous works \cite{carion2020end,strudel2021segmenter}, the information of $K$ classes will be encoded as the input for the decoder. Each class $k$ will be represented by a learnable vector with the same shape as the image patch. The token list for all classes can be denoted as $z_0 \in \mathbb{R}^{K \times D}$. As shown Figure \ref{overview}, $z_0$ will be concatenated with patch sequence $s_L$ (denote as $s_{L,0}$ for decoder) to form the input for decoder as $[s_{L,0}, z_0] \in \mathbb{R}^{(N+K) \times D}$.

After $M$ layers of the transformer (same structure as encoder), the output of the patch-class combination is denoted as $[s_{L,M}, z_M]$. $z_M$ is considered as the weight for each class and will be multiplied with $s_{L,M}$ for the mask prediction $\hat{y}$ as follows:
\begin{align}
    \hat{y} = s_{L,M} \cdot z_M^T,
\end{align}
where $\cdot$ is the matrix multiplication and $\hat{y} \in \mathbb{R}^{H \times W \times K}$. We will further utilize $\hat{y}$ as a preliminary estimation of the window instance and then modify $\hat{y}$ using our pre-set priory knowledge.

\subsection{Line Acquisition, Filtering, and Revision}\label{LAFR}
Different from previous works that adopt heavy and rigid deep learning-based object detection models for the improvement of facade prediction. In this research, we want to design a simple yet efficient method named LAFR. We take advantage of facade priory, that is, most of the facade elements (e.g., windows, doors) are generally quadrilateral. Based on this, we adopt traditional line detection methods to realize localization and further modify the facade prediction.

\begin{figure*}[!t]
	\centering
	\includegraphics[width=0.72\linewidth]{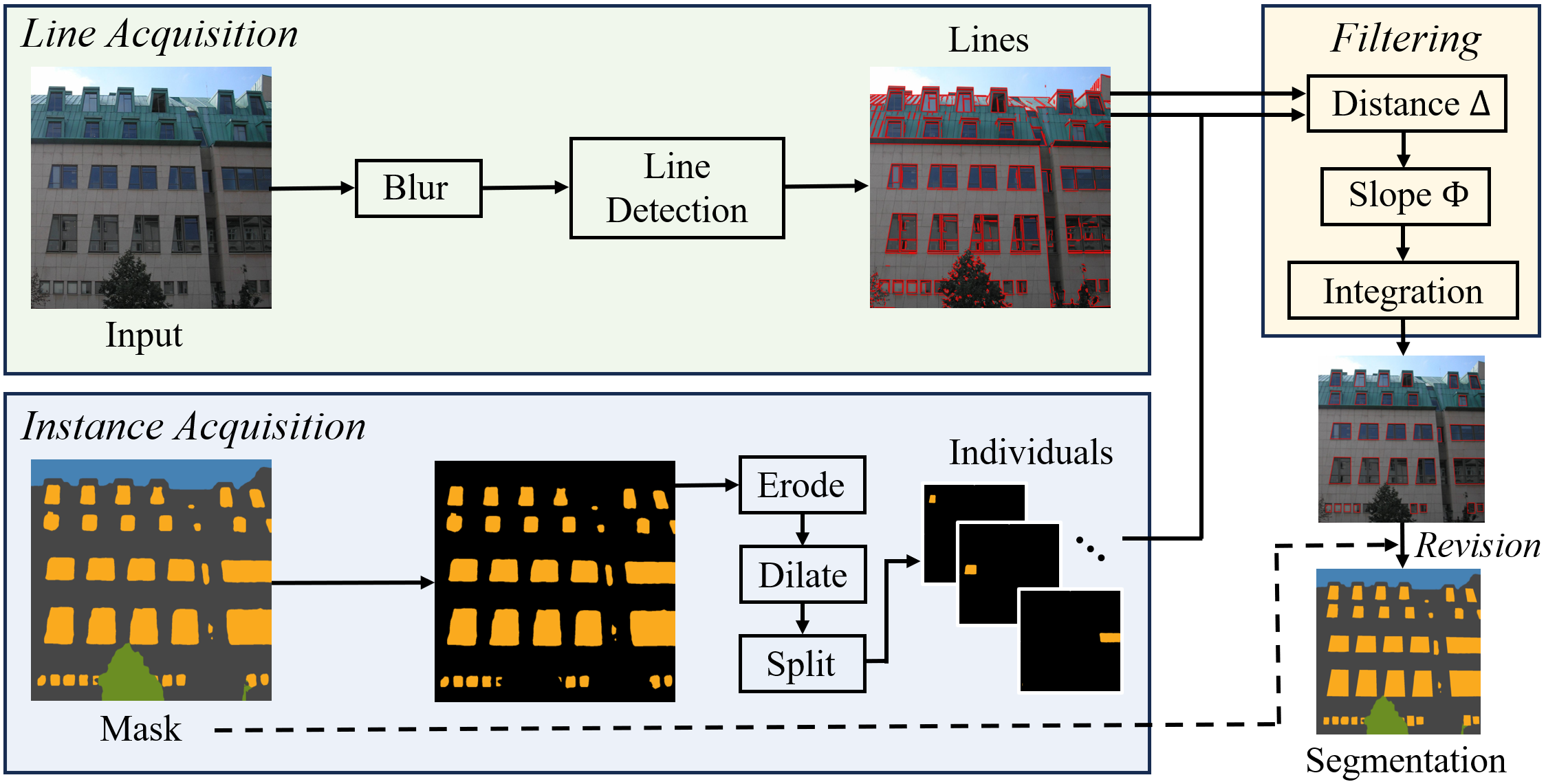}
	\caption{Our proposed Line Acquisition, Filtering, and Revision (LAFR) algorithm. It acquires lines from the input image and window instance from the predicted mask. After filtering, we use the integrated outline to revise the predicted mask.}
	\label{Filtering}
\end{figure*}

\subsubsection{Line Acquisition}
There many methods designed for line detection, e.g., Hough and Line Segment Detector (LSD). We adopted LSD for our target since it is a powerful method for detecting line segments in digital images. It operates by analyzing the image's edge map to identify candidate line segments. LSD employs a series of steps, including the computation of region orientations, pixel grouping, and line validation. By considering both local and global information, the LSD algorithm effectively detects line segments with high accuracy and robustness. Its ability to handle complex scenes, varying lighting conditions, and noisy images makes it a popular choice in computer vision applications such as object recognition, image stitching, and scene analysis.

In our revision pipeline, depicted in Figure \ref{Filtering}, we employ a series of steps to enhance the accuracy of our line detection. Initially, the input image undergoes dual Gaussian blurring processes, effectively eliminating extremely short line segments caused by noise. The size of convolution kernels is $5 \times 5$ and $3 \times 3$, respectively. The standard deviation of the Gaussian function is 5. Subsequently, the LSD algorithm is employed to detect a set of $J$ potential line segments, denoted as a collection $G = \{(\alpha_{1j}, \beta_{1j}, \alpha_{2j}, \beta_{2j}) \mid j=1,\dots, J\}$. These line segments are characterized by their quaternion representations, which record the coordinates of their start and end points.

\subsubsection{Instance Acquisition}
Our objective is to identify each facade element as our target on the facade revision and incorporate prior knowledge to $G$ to refine their prediction. Using the ViT model, we can generate a preliminary prediction mask $\hat{y}$, which provides a preliminary estimate. To isolate each individual facade element, we retain only the pixels within the region predicted as a facade element. Figure \ref{Filtering} illustrates the subsequent steps in our process, where we employ erosion and dilation techniques to address noise and subtle connections between windows in the prediction. Then, we calculate the connected components for each element and split them into $B$ individual masks demoted as $F = \{(\hat{\alpha}_{1b}, \hat{\beta}_{1b}, \hat{\alpha}_{2b}, \hat{\beta}_{2b}) \mid b=1,\dots, B\}$. Each $b$ is also a quaternion, where the first and last two values are the coordinate of the minimum external rectangle $R = [r_{top}, r_{bottom}, r_{left}, r_{right}]$ (top, left, bottom, and right edge) for the mask, respectively. 

\subsubsection{Filtering and Revision}
We manually define some policies to filter line segments $G$, transforming lines into a quad that can perfectly frame out each facade element. We set the minimum external rectangle from each $b$ as the anchor, and the filtering is organized as follows:

\begin{equation}
    E = \oint_{b}^B \Psi (\oint_{j}^J \Xi [\Delta (b, j, r); \Theta (b, j, r)]),
\end{equation}

where $E \in \mathbb{R}^{B \times R}$ serves as a record, indicating the assigned line segments to each anchor's edge. The circular pair calculation between $G$ and $F$ is denoted by $\oint$. The function $\Xi[\cdot]$ computes each $j$ using four anchor edges, employing two threshold functions, $\Delta(\cdot)$ and $\Theta(\cdot)$, with threshold values $\delta$ (default as 20) and $\theta$ (default as 0.1), respectively. These functions are responsible for distance and angle calculations. Line segments far from the anchor are discarded, while the remaining segments are assigned to their respective edges based on the angle. It is important to note that each edge can only accommodate a single line segment with the smallest distance (we only keep the longest line among all qualified candidates). In cases where no satisfactory line segment is found for an edge, a blank value is recorded.

The function $\Psi(\cdot)$ is responsible for integrating the four line segments associated with each edge of the anchor, resulting in the formation of a new external rectangle (shown in Figure \ref{Filtering}). However, if an anchor exhibits a blank edge, we discard it for further consideration. This is due to the inability to accurately determine the facade element's shape in such cases. As a result, the record for this particular anchor is deemed invalid and is subsequently skipped in the subsequent stages of the revision process.

For the revision, we directly use $E$ to replace the window region in $\hat{y}$ correspondingly. Thus, the final mask prediction for the facade is $y$.

\section{Results}\label{sec:results}
\subsection{Datasets and Metrics}
We compared our method with other SOTA methods in two open-source datasets and our CFP dataset (statistics of object number are showing Table \ref{tab_quan_dataset}). There are also four metrics adopted for quantification.
\paragraph{Our CFP Dataset} CFP represents a meticulously curated collection of street view images acquired through state-of-the-art equipment, resulting in a total of 602 high-resolution images, each boasting a 2560x1440 pixel resolution. These images were captured across six cities: Osaka, Tokyo, Toronto, Shanghai, Nanjing, and Nanchang, providing a diverse and comprehensive view of urban landscapes from various angles. Within the CFP dataset, we have meticulously categorized images into nine distinct classes: buildings, windows, doors, roofs, trees, sky, people, cars, and signs. This comprehensive categorization enables detailed scene understanding and segmentation, allowing for a wide range of applications. The dataset encompasses a rich variety of street views, encompassing the charm of residential neighborhoods and the dynamic energy of bustling city centers. To ensure the highest quality and accuracy, each image is meticulously annotated at the instance mask level. For this labeling process, we enlisted the expertise of two dedicated annotators, guaranteeing the precision of our dataset. 80\% of the data are used for training, 10\% for validation, and 10\% for testing. As a result, CFP is an invaluable resource for many applications. Researchers and practitioners alike can leverage this dataset to gain deep insights into urban environments, enhance their algorithms, and address real-world challenges related to facade parsing, scene understanding, and more. 

\paragraph{ECP 2011 \cite{teboul2010ecole}} The ECP dataset comprises 104 rectified images showcasing facades characterized by Haussmannian-style architecture. Due to the inherent imprecision in the original annotations, we have adopted the annotations established in prior research \cite{ma2020pyramid}. These annotations classify elements into eight distinct categories: windows, walls, balconies, doors, roofs, chimneys, skies, and shops.

\paragraph{RueMonge 2014 \cite{riemenschneider2014learning}} This dataset has 428 high-resolution and multiview images from Paris streets. It was designed to support developing and evaluating algorithms and models for various urban scene understanding tasks, such as visual localization, 3D reconstruction, and visual odometry.  It provides annotations with seven semantic categories for 2D images, including door, shop, balcony, window, wall, roof, and sky. It is worth noting that there are actually only 32 unique facade images in this dataset. Thus, the number of each class is far less than we statistics in Table \ref{tab_quan_dataset}. 

\paragraph{Metrics} We evaluate the segmentation results using pixel-level accuracy (Acc), the average accuracy of prediction for each category (Class avg.), F1-score for quantifying imbalanced classes, and mean intersection over union (mIoU).
 Here, we denote $\text{hist}$ as the confusion matrix for pixel-level classification. The definition for $Acc$ is as follows:

\begin{equation}
Acc = \frac{{{TP} + {TN}}}{{{TP} + {TN} + {FP} + {FN}}},
\end{equation}
where $TP$, $TN$, $FP$, and $FN$ are the number of true positive, true negative, false positive, and false negative. The definition of F1-score is as follows: 

\begin{align}
Precision = \frac{TP}{TP + FP}, \\
Recall = \frac{TP}{TP + FN}, \\
F1-score = 2 * \frac{Precision * Recall}{Precision + Recall}.
\end{align}
It is used to assess the accuracy and robustness of the model's segmentation results, especially when dealing with imbalanced class distributions.

Class avg. is calculated to quantify per-class accuracy as follows:
\begin{equation}
Class\_avg= \frac{1}{K}\sum_{i=1}^{K}\frac{diag(hist)[i]}{\sum(hist[:, i])},
\end{equation}
where $diag(\cdot)$ is diagonal value.

\begin{table}[t]
\caption{Number of objects in three datasets.}
\centering
\resizebox{0.70\columnwidth}{!}{
\begin{tabular}{lccc}
\toprule
class & ECP2011 & RueMonge 2014 & CFP\\
\midrule
building & 104 & 32 & 1545\\
window & 2976 & 8416 & 12048\\
balcony & 1991 & 2620 & -\\
wall & 104 & 428 & -\\
door & 94 & 311 & 896\\
roof & 104 & 428 & 383\\
shop & 198 & 642 & -\\
tree & - & - & 611\\
sky & - & 402 & 720\\
people & - & - & 446\\
car & - & - & 531\\
sign & - & - & 923\\
\midrule
total & 5580 & 13279 & 18103\\
\bottomrule
\end{tabular}
}
\label{tab_quan_dataset}
\end{table}

(\text{mIoU}) is computed as the average of the class-wise IoU:
\begin{equation}
mIoU= \frac{{1}}{{N}} \sum_{i=1}^{N} {IoU}_i,
\end{equation}

where $N$ is the number of classes and $\text{{IoU}}_i$ is the IoU value for class $i$.

\begin{table*}[t]
\caption{Segmentation results on proposed CFP dataset. We compare the performance of vanilla segmentation models and models designed for facade parsing.}
\centering
\resizebox{0.60\textwidth}{!}{
\begin{tabular}{llcccc}
\toprule
Method & Backbone & Acc & Class\_avg & F1-score & mIoU\\
\midrule
PSPNet \cite{zhao2017pyramid} & ResNet-101 & 88.32 & 78.01 & 78.47 & 60.03\\
Deeplabv3+ \cite{chen2017rethinking} & ResNeSt-101 & 88.48 & 78.33 & 78.90 & 60.89\\
OCR \cite{yuan2020object} & HRNetV2 & 87.85 & 76.95 & 77.42 & 58.02\\
Swin-L UperNet \cite{liu2021swin} & Swin-B/16 & 87.95 & 77.13 & 77.58 & 58.65\\
SETR-L MLA \cite{zheng2021rethinking} & ViT-B/16 & 88.30 & 77.96 & 78.44 & 60.14\\
Segmenter \cite{strudel2021segmenter} & ViT-B/16 & 88.72 & 79.35 & 79.78 & 61.60\\
\midrule
DeepFacade \cite{liu2017deepfacade} & FCN+Faster-RCNN & 88.47 & 78.30 & 78.81 & 60.85\\
Femiani \textit{et al.} \cite{femiani2018facade} & AlexNet & 86.19 & 70.51 & 71.45 & 50.22\\
ALKNet \cite{ma2020pyramid} & ResNet-FCN & 88.76 & 79.38 & 79.86 & 61.74\\
FacMagNet \cite{DAI2021107921} & FCN+Mask-RCNN & 88.41 & 78.25 & 78.94 & 60.62\\
\midrule
RTFP \textit{(Ours)} &  ViT-B/16+LAFR & \textbf{88.80} & \textbf{79.75} & \textbf{80.63} & \textbf{61.95} \\
RTFP \textit{(Ours)} & ViT-L/16+LAFR & 88.78 & 79.47 & 80.06 & 61.87 \\
\bottomrule
\end{tabular}
}
\label{acc_tab1}
\end{table*}

\begin{table}[t]
\caption{Segmentation results on ECP 2011 dataset.}
\centering
\resizebox{1\columnwidth}{!}{
\begin{tabular}{lcccc}
\toprule
Method & Acc & Class\_avg & F1-score & mIoU \\
\midrule
PSPNet \cite{zhao2017pyramid} & 93.62 & 90.95 & 91.08 & 83.76 \\
DANet \cite{fu2019dual} & 93.70 & 91.04 & 91.21 & 84.02 \\
Deeplabv3+ \cite{chen2017rethinking} & 93.75 & 91.20 & 91.34 & 84.26 \\
Segmenter \cite{strudel2021segmenter} & 93.82 & 91.63 & 91.81 & 84.68 \\
\midrule
Femiani \textit{et al.} \cite{femiani2018facade} & 82.79 & 79.06 & 79.21 & 72.25 \\
Rahmani \textit{et al.} \cite{rahmani2018high} & 92.20 & 91.00 & - & - \\ 
DeepFacade \cite{liu2017deepfacade} & 93.86 & 91.75 & 91.86 & 84.75 \\
ALKNet \cite{ma2020pyramid} & 93.88 & 91.80 & 91.98 & 84.81 \\
\midrule
RTFP ViT-B/16 \textit{(Ours)} & \textbf{93.92} & \textbf{91.88} & \textbf{92.13} & \textbf{84.93} \\
\bottomrule
\end{tabular}
}
\label{acc_ECP}
\end{table}

\begin{table}[t]
\caption{Segmentation results on RueMonge 2014 dataset.}
\centering
\resizebox{1\columnwidth}{!}{
\begin{tabular}{lcccc}
\toprule
Method & Acc & Class\_avg & F1-score & mIoU \\
\midrule
PSPNet \cite{zhao2017pyramid} & 87.20 & 80.31 & 81.26 & 70.94 \\
Deeplabv3+ \cite{chen2017rethinking} & 87.80 & 81.02 & 82.50 & 72.92 \\
Segmenter \cite{strudel2021segmenter} & 87.98 & 82.62 & 83.79 & 73.10 \\
\midrule
DeepFacade \cite{liu2017deepfacade} & 87.99 & 82.70 & 83.84 & 73.13 \\
ALKNet \cite{ma2020pyramid} & 88.01 & 82.75 & 83.90 & 73.21 \\
\midrule
RTFP ViT-B/16 \textit{(Ours)} & \textbf{88.12} & \textbf{83.14} & \textbf{84.17} & \textbf{73.46} \\
\bottomrule
\end{tabular}
}
\label{acc_RueMonge}
\end{table}

\begin{table}[t]
\caption{Computation cost comparison to previous facade segmentation methods using CPU/GPU.}
\centering
\resizebox{0.92\columnwidth}{!}{
\begin{tabular}{lccc}
\toprule
Device & FacMagNet \cite{DAI2021107921} & DeepFacade \cite{liu2017deepfacade} & RTFP \\
\midrule
CPU & 5.62 s & 6.25 s & \textbf{1.97 s} \\
GPU & 1.03 s & 1.44 s & - \\
\bottomrule
\end{tabular}
}
\label{efficeincy}
\end{table}

\subsection{Experimental Settings}
\textbf{ViT Model Structure} The structure of our encoder follows \cite{strudel2021segmenter} and we tried the "Tiny", "Small", "Base", and "Large" model sizes (performance is described in Section \ref{ablation}). We also let the models use different resolutions of input image corresponding to patch sizes $8 \times 8$, $16 \times 16$, and $32 \times 32$. For the decoder, we adopt only two layers ($M=2$) of transformers with a head number of 8. 

\textbf{Model Pre-training} Previous works adopt ImageNet \cite{russakovsky2015imagenet} initialization for easier training. However, the data distribution between ImageNet and facade datasets is quite different. The parameters pre-trained on ImageNet are unsuitable for the initialization of the ViT model in the facade segmentation task. We thus adopt an MAE \cite{he2022masked} pre-training to initialize ViT methods. As MAE is a self-supervised method, we use all three datasets as well as Cityscapes \cite{cordts2016cityscapes} as external data for the pre-training process. MAE is implemented based on its official training pipeline. For all the CNNs-based methods, we use ImageNet initialization.

\textbf{Optimizer Setting} We used cross entropy as the loss function and AdamW \cite{kingma2014adam} as the optimizer. Following the experimental setting of DeepLab \cite{chen2017rethinking}, we adopt the "poly" learning rate decay. The learning rate is reduced gradually over time during training and starts from 0.0001. For all the settings, we implement 60 epochs. Our experiments are implemented on a single A6000 GPU with an Intel(R) Xeon(R) W-2255 CPU.

\textbf{Augmentation} We adopt mean subtraction, a ratio between 0.5 and 2.0 for the random resizing of the image, and random flipping from left to right. We randomly crop images to a fixed size of $640 \times 640$ for the proposed dataset and $448 \times 448$ for other datasets. During inference, we adopt dense cropping over the whole image \cite{chen2017rethinking}.

\begin{table*}[t]
\caption{Segmentation results of each class on our CFP dataset. All results are evaluated by mIoU.}
\centering
\resizebox{0.91\textwidth}{!}{
\begin{tabular}{llccccccccc}
\toprule
Method & building & window & sky & roof & door & tree & people & car & sign & mIoU\\
\midrule
PSPNet \cite{zhao2017pyramid} & 83.44 & 62.47 & 92.45 & 48.81 & 47.53 & 52.55 & 41.47 & 76.71 & 35.02 & 60.03\\
Segmenter \cite{strudel2021segmenter} & 85.28 & 63.34 & 93.20 & 54.72 & 54.45 & 52.61 & 41.47 & 78.03 & 31.58 & 61.60\\
\midrule
Femiani \textit{et al.} \cite{femiani2018facade} & 80.93 & 59.81 & 90.66 & 38.52 & 23.37 & 51.50 & 08.17 & 73.74 & 25.26 & 50.22 \\
DeepFacade \cite{liu2017deepfacade} & 85.71 & 64.68 & 93.25 & 55.74 & 53.58 & 52.59 & 41.43 & 78.02 & 31.64 & 61.85\\
FacMagNet \cite{DAI2021107921} & 84.81 & 65.19 & 92.25 & 52.98 & 54.07 & 53.24 & 34.29 & 77.87 & 30.88 & 60.62\\
ALKNet \cite{ma2020pyramid} & 83.36 & 62.54 & 92.47 & 48.82 & 46.75 & 54.13 & 28.88 & 76.79 & 36.00 & 58.86\\
\midrule
RTFP ViT-B/16 \textit{(Ours)} & \textbf{85.49} & \textbf{65.32} & 93.15 & 54.72 & \textbf{54.73} & 52.86 & 41.47 & 77.96 & 31.81 & \textbf{61.95}\\
\bottomrule
\end{tabular}
}
\label{acc_tab2}
\end{table*}

\subsection{Comparison to Previous Works}
In this section, we compared our method to SOTA methods in semantic segmentation tasks (PSPNet \cite{zhao2017pyramid}, Deeplabv3+ \cite{chen2017rethinking}, OCR \cite{yuan2020object}, Swin-L UperNet \cite{liu2021swin}, SETR-L MLA \cite{zheng2021rethinking}, and Segmentor \cite{strudel2021segmenter}) and methods specially designed for facade segmentation tasks (DeepFacade \cite{liu2017deepfacade}, FacMagNet \cite{DAI2021107921}, and ALKNet \cite{ma2020pyramid}). For CFP, our revision methods are only related to windows and doors (other classes are rare quadrilateral). For ECP 2011 and RueMonge 2014, revision is implemented for all classes besides building and sky. We aim to showcase the effectiveness and accuracy of our method by evaluating its performance against these prominent baselines. The following discussion outlines the evaluation of three datasets used for the comparison, providing a thorough analysis of the results obtained. 

\begin{figure*}[!t]
	\centering
	\includegraphics[width=0.92\textwidth]{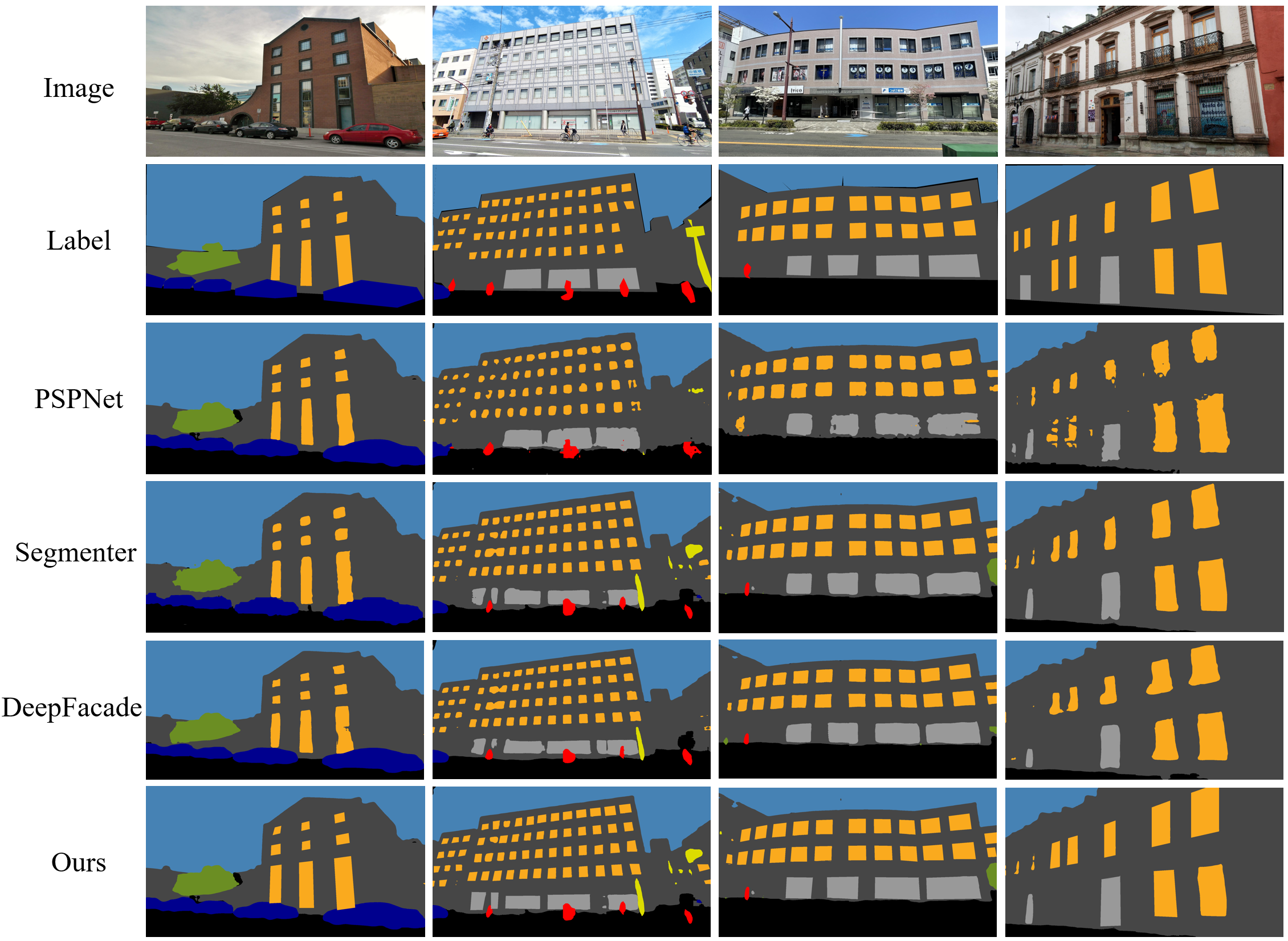}
	\caption{Qualitative comparison to previous segmentation methods on our CFP dataset.}
	\label{result_compare}
\end{figure*}

As presented in Table \ref{acc_tab1}, we first want to showcase the experimental results obtained from our CFP dataset. Notably, our RTFP consistently outperforms all competitive methods, with the most exceptional performance emerging when utilizing ViT-Base with a patch size of 16 (for more comprehensive insights, refer to section \ref{ablation}). The comparison also indicates that ViT-based approaches exhibit superior performance compared to their CNN-based counterparts. For instance, the mIoU of Segmentor, a ViT method, is marginally superior by approximately 0.71\% than CNN-based Deeplabv3+. This observation substantiates our hypothesis that the ViT architecture holds promise in the context of facade segmentation tasks. Its holistic perception capability aligns well with the requirements of comprehending complex facade scenes, ultimately contributing to improved segmentation outcomes. We can also find that a heavy extra revision (e.g., DeepFacade and FacMagNet) will get better results. We further demonstrate the performance comparison over ECP and RueMonge (Table \ref{acc_ECP} and Table \ref{acc_RueMonge}), respectively. RTFP still shows its superiority over previous works.

Furthermore, we provide an in-depth analysis of the IoU scores for each individual class, as outlined in Table \ref{acc_tab2}. Notably, beyond attaining superior IoU performance compared to prior methods, the distinctive advantage of our LAFR algorithm is also prominently apparent upon scrutinizing these outcomes. Since we employ Segmentor as our ViT structure for preliminary segmentation, compared to the results of raw Segmentor, the incremental IoU improvements facilitated by the LAFR algorithm are evident in the building, window, and door classes, with enhancements of approximately 0.21\%, 1.98\%, and 0.28\%, respectively. This compelling evidence underscores LAFR's capacity to refine preliminary segmentation outputs precisely. It is pertinent to mention that the non-training of the LAFR algorithm extends to its applicability across various segmentation models. We illustrate this compatibility in the \ref{ablation} section, showcasing its ability to integrate with diverse segmentation architectures.

\begin{figure*}[!t]
	\centering
	\includegraphics[width=1.\textwidth]{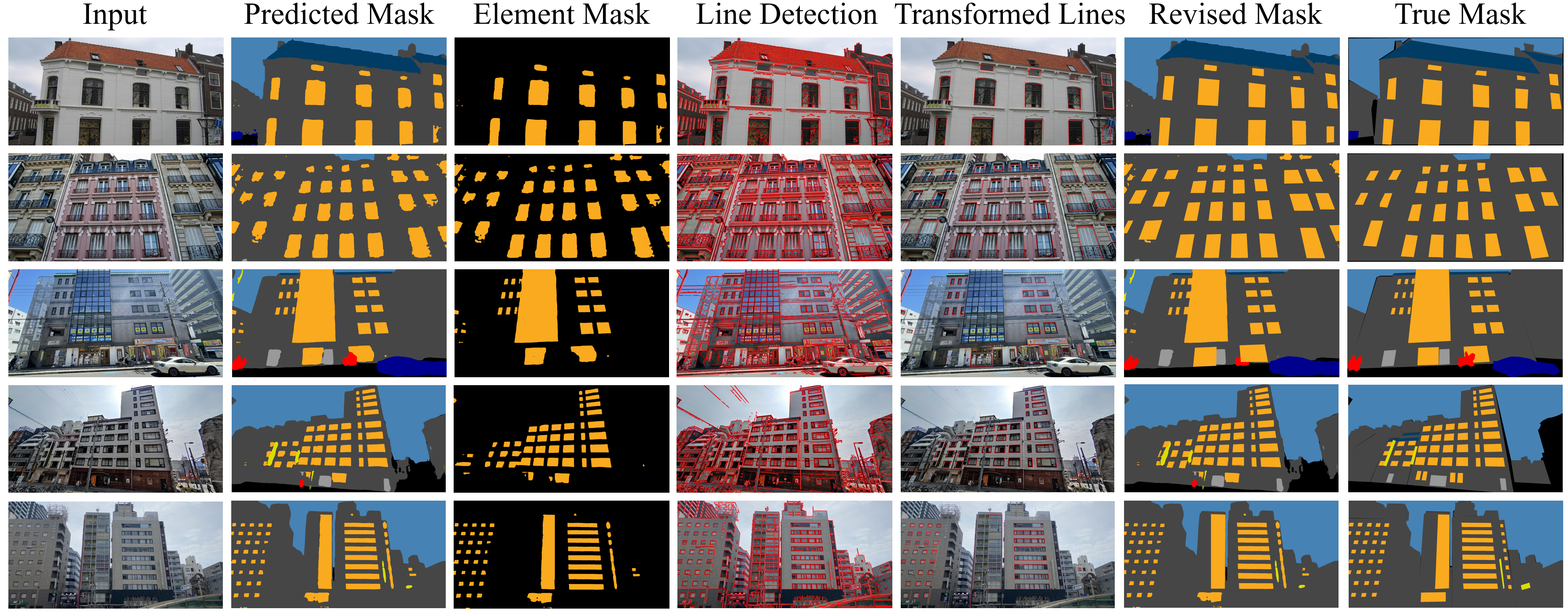}
	\caption{Inference samples of our LAFR pipeline. From left to right are the input image, the predicted mask generated by the segmentation model, the facade element mask (for windows), all detected lines using the LSD method, transformed lines after undergoing the LAFR process, the revised mask, and the ground truth.}
	\label{revision}
\end{figure*}

In Figure \ref{result_compare}, we systematically compare the qualitative results of our RTFP model and several competing approaches. Notably, the outcome achieved by the ViT-based Segmentor surpasses that of the CNNs-based PSPNet in terms of coherence and delineation clarity. Additionally, the ViT-based Segmentor excels in predicting classes with limited instance samples, a prime example being people and cars. Despite these advancements, it's important to highlight an observable limitation. The precision of window outlines remains somewhat untidy, thus inevitably affecting the overall accuracy of building predictions. For the method designed for facade revision, DeepFacade directly used the bounding box detection results from Faster-RCNN to revise the facade prediction. It may work well for the rectified scene, while most of the images are taken from different angles for our CFP dataset. Our LAFR algorithm uses the line segment rules of the facade element itself to generate a conformable outline, which greatly improves the accuracy of correction. It is obvious that our segmentation results show a straighter shape for windows and doors. However, LAFR does not succeed for all facade instances. We can find a few samples that fail to be revised.

Efficiency stands as a distinct advantage of RTFP over previous facade segmentation methods reliant on object detection. Traditionally, object detection models necessitate substantial GPU memory for computation, leading to time-intensive processes. Consequently, this approach may not align with the cost-effectiveness demanded by real-world applications. In Table \ref{efficeincy}, we undertake a comprehensive comparison of RTFP's computational efficiency against DeepFacade (utilizing Faster-RCNN) and FacMagNet (employing Mask-RCNN). This analysis specifically focuses on the computational costs attributed to the revision module, omitting the computational load posed by the backbones. The results demonstrate that our proposed LAFR algorithm outpaces the GPU configurations of both DeepFacade and FacMagNet by a noticeable margin (4.28 s and 3.65 s faster). Remarkably, even when constrained to CPU resources alone, RTFP's computational costs remain competitive. This attests to its potential suitability for real-world applications, unburdened by device limitations.

\subsection{Revision Demonstration}

In Figure \ref{revision}, we present a series of samples illustrating the performance of our LAFR pipeline. The images are arranged from left to right, showcasing the following components: the input image, the predicted mask generated by the segmentation model, the facade element mask, all detected lines using the LSD method, transformed lines after undergoing the LAFR process, the revised mask, and the ground truth. A noteworthy observation is the irregular outline of windows in the predicted mask produced by the segmentation model, highlighting a limitation of conventional segmentation models. This irregularity is particularly evident in the window mask displayed in the third row. In the fourth row, we display the results of line detection using the LSD algorithm. Notably, LSD detects a substantial number of line segments, many of which closely adhere to the edges of buildings and windows. This observation substantiates our hypothesis that employing a straightforward line segment detection approach can yield precise window positioning. However, it is in the fifth column, where the line segments have been transformed through the LAFR algorithm, that we witness a marked improvement. These integrated line segments accurately delineate the windows, demonstrating the potential for revising the original prediction mask in subsequent iterations of our pipeline.

LAFR exhibits strong performance across the majority of scenarios within our dataset. However, its effectiveness is notably influenced by the quality of the initial predicted mask. Predicted masks are relatively high quality in the first to third rows. Based on such prior, LAFR excels in providing valuable revision guidance. Conversely, the last two examples present a challenge. These instances involve numerous window elements, and the predictions for windows may be either incorrect or incomplete. In such situations, LAFR faces the possibility of either refraining from revising certain windows or making erroneous revisions. The dependence on segmentation model results represents a limitation of our current LAFR implementation. This phenomenon is also explored in the compatibility experiments in Table \ref{acc_compat}. We acknowledge this limitation and are committed to addressing it in our future research endeavors, striving to enhance the robustness and reliability of our revision algorithm.

\subsection{Ablation Study} \label{ablation}

\begin{table}[t]
\caption{Experiments for compatibility evaluation.}
\centering
\resizebox{0.75\columnwidth}{!}{
\begin{tabular}{lcccc}
\toprule
Segmentation Model & Acc & mIoU \\
\midrule
UNet \cite{ronneberger2015u} & - 00.42 & - 01.91 \\
FCN \cite{long2015fully} & - 00.25 & - 01.34 \\
PSPNet \cite{zhao2017pyramid} & + 00.02 & + 00.10 \\
Deeplabv3+ \cite{chen2017rethinking} & + 00.07 & + 00.27 \\
Segmenter \cite{strudel2021segmenter} & + 00.08 & + 00.35 \\
\bottomrule
\end{tabular}
}
\label{acc_compat}
\end{table}

\begin{table}[t]
\caption{The segmentation performance using different pre-training. All experiments use our RTFP as the default setting and implement the same training process.}
\centering
\resizebox{0.9\columnwidth}{!}{
\begin{tabular}{lcccccccc}
\toprule
&\multicolumn{2}{c}{ECP2011} &\multicolumn{2}{c}{RueMonge} &\multicolumn{2}{c}{CFP}\\
\cmidrule(lr){2-3} \cmidrule(lr){4-5} \cmidrule(lr){6-7}
Pre-training & Acc  & mIoU & Acc & mIoU & Acc & mIoU \\ 
\midrule
Random & 29.27  & 12.11 & 23.55 & 08.17 & 32.50 & 19.24\\
ImageNet & 93.55  & 83.49  & 87.76 & 72.27 & 88.39 & 59.41 \\ 
MAE & \textbf{93.92}  & \textbf{84.93} & \textbf{88.12} & \textbf{73.46} & \textbf{88.80} & \textbf{61.95} \\
\bottomrule
\end{tabular}}
\label{pre-train}
\end{table}

\textbf{Compatibility} Our LAFR algorithm offers a straightforward approach to revision and is adaptable to various model architectures beyond just the Segmentor \cite{strudel2021segmenter}. In Table \ref{acc_compat}, we extend the application of LAFR to other models and analyze its impact. Notably, we observe a marked enhancement in performance for PSPNet \cite{zhao2017pyramid} and Deeplabv3+ \cite{strudel2021segmenter}. Conversely, a performance decrement is evident when applying LAFR to UNet \cite{ronneberger2015u} and FCN \cite{long2015fully}. As discussed in the methodology section \ref{LAFR}, LAFR relies on the quality of the initial segmentation results. UNet and FCN, being early pioneers in semantic segmentation, may exhibit reduced accuracy on our dataset due to their older design paradigms. In summation, our findings suggest that the efficacy of LAFR is notably bolstered by the utilization of advanced segmentation models, underscoring the importance of employing state-of-the-art architectures for optimal performance.

\begin{figure}[!t]
	\centering
	\includegraphics[width=0.85\columnwidth]{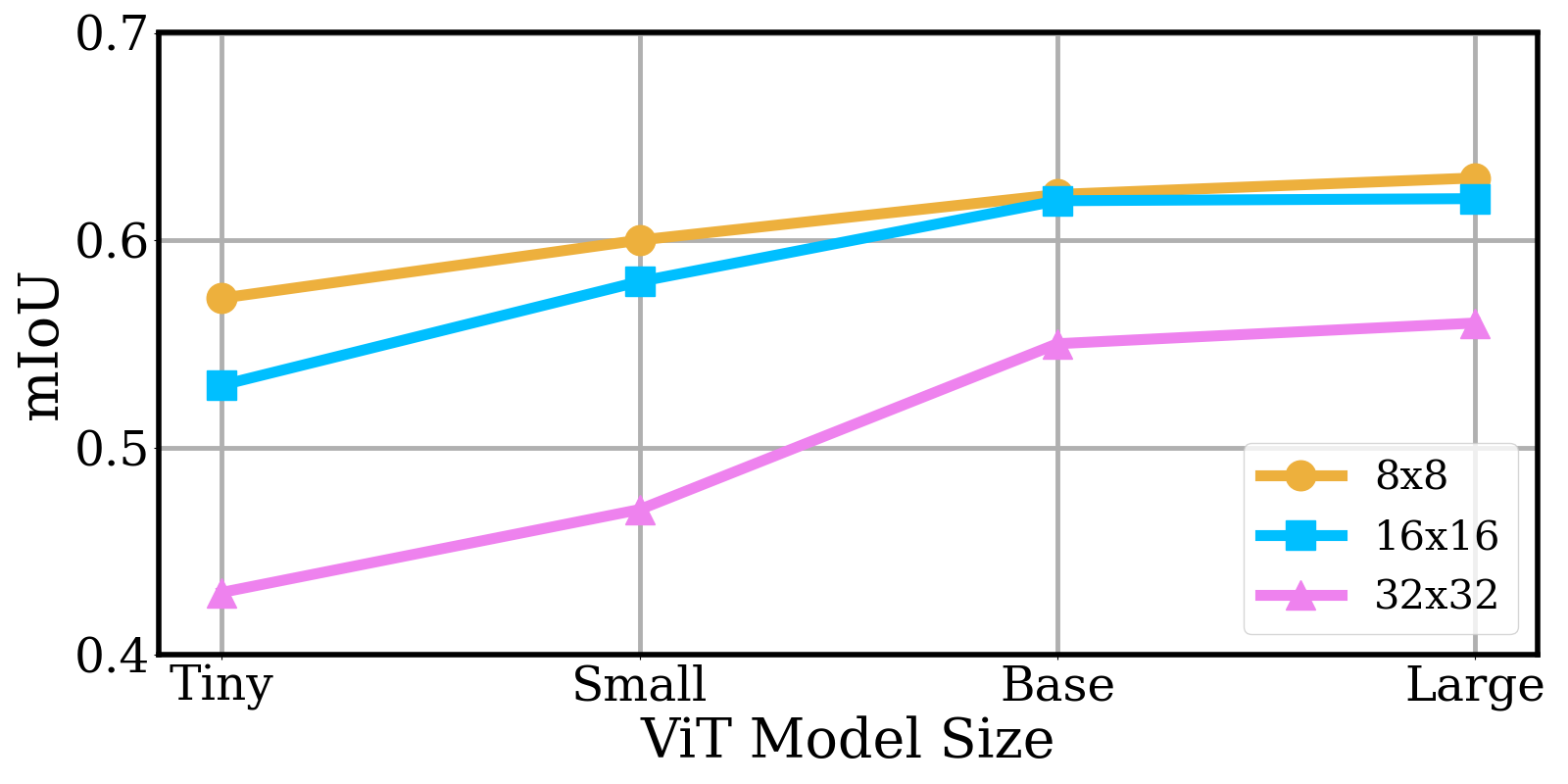}
	\caption{Ablation experiments over ViT model structure and patch size (quantified with mIoU).}
	\label{abalation_vit}
\end{figure}

\textbf{Pre-training} A key focus of our research is to investigate the impact of pre-trained models on the efficacy of facade segmentation tasks. To ensure a fair and consistent evaluation, all experiments adopt our RTFP as the default configuration and adhere to an identical training protocol. As illustrated in Table \ref{pre-train}, it becomes evident that MAE-based pre-training consistently outperforms other methods across all three datasets. Notably, there is a substantial performance gap when compared to models pre-trained on ImageNet, which lags behind. Conversely, models initialized with random weights yield notably inferior results. These findings provide robust evidence of the effectiveness of MAE-based pre-training for enhancing the performance of models in facade segmentation tasks

\textbf{ViT Structure} The configuration of the ViT model plays a pivotal role in influencing the performance of our RTFP. To investigate its impact, we conducted an experiment focusing on two crucial aspects: the model size and patch size. As depicted in Figure \ref{abalation_vit}, our findings reveal a clear trend: both increasing the ViT model's size and reducing the patch size positively influence prediction accuracy. Nevertheless, it's worth noting that the improvement from the "Base" model to the "large" model appears to be relatively marginal. However, substantial computational demands are posed by larger ViT models, especially when dealing with smaller patches. In light of these observations, we recommend the utilization of the "Base" ViT model with a patch size of 16. This configuration strikes a practical balance between prediction performance and computational efficiency, making it an optimal choice for the RTFP system.


\section{Conclusion}\label{sec:conclusion}
In this paper, we released a new dataset named CFP that serves as the benchmark for facade parsing. The creation of this dataset involved meticulous data collection and annotation, ensuring its high quality and relevance to real-world scenarios. Previous works have primarily focused on offering datasets comprising simplistic single-family building facade images, while our dataset takes a more comprehensive data source. Recognizing the intricate demands of real-world applications, our collection spans a diverse range of building facade images, ranging from straightforward residential contexts to the intricacies of densely populated urban areas. Furthermore, our dataset encompasses images of buildings captured from various angles, providing a richer and more comprehensive representation of architectural diversity in urban environments. We aim to foster collaboration and facilitate fair comparisons between different methods by offering a common dataset for evaluation. This standardized benchmark will accelerate progress and promote transparency and reproducibility in facade parsing research. We believe that CFP will significantly contribute to advancing state-of-the-art facade parsing, providing researchers and practitioners with a valuable resource for evaluating and comparing their algorithms. 

We also proposed a new facade parsing pipeline RTFP based on vision transformers and line integration. Our empirical findings underscore the remarkable advantages of employing ViT. Notably, the incorporation of the pre-training method, MAE, amplifies the prowess of ViT even further. These results are indicative of ViT's immense potential for application across a spectrum of facade parsing scenarios. With its inherent capability to capture comprehensive global context within images, we envision ViT as a versatile tool for a wide array of facade parsing applications. ViT exhibits a unique proficiency in discerning intricate relationships among distant elements within a facade—a pivotal factor for achieving accurate parsing of complex architectural structures. This, in turn, facilitates finer-grained segmentation and a deeper understanding of the constituent components comprising facades. In addition, an efficient yet accurate revision method LARF is designed to improve the segmentation results further. Leveraging prior knowledge of building facades, we demonstrate that simple line segment detection and integration can also match or exceed additional object detection models. However, our method currently only works on base facade elements (e.g., windows and doors in our CFP) and relies on the results of the segmentation model. Our future research will aim to improve these shortcomings.

As we conclude this work, we are poised to continue exploring new frontiers in facade parsing, addressing existing limitations, and extending the applicability of these techniques to a broader array of architectural elements. We remain committed to advancing the field, pushing the boundaries of what is possible, and contributing to the ever-evolving landscape of computer vision and architectural analysis.

\section*{Acknowledgment}
This work is partly supported by JSPS KAKENHI Grant Number 19K10662, 20K23343, 21K17764, and 22H03353. We also appreciate Tong Zhao and Yuting Wang for their annotation of our dataset during this research.

\bibliographystyle{elsarticle-harv} 
\bibliography{example}

\end{document}